%% file: acl_latex.tex
\pdfoutput=1

\documentclass[11pt]{article}

\usepackage[]{acl}

\usepackage{times}
\usepackage{latexsym}

\usepackage[T1]{fontenc}

\usepackage[utf8]{inputenc}

\usepackage{microtype}

\usepackage{inconsolata}
\usepackage{makecell}
\usepackage{float}
\usepackage{bm}
\usepackage{booktabs}
\usepackage{multirow}
\usepackage{xcolor}
\usepackage{colortbl}
\usepackage{xspace}
\usepackage{subfig}
\usepackage{graphicx} 
\usepackage{color}
\usepackage{paralist}
\usepackage{acronym}
\usepackage{tabularx}
\usepackage[ruled,linesnumbered]{algorithm2e}

\usepackage{amsmath,amsfonts,amsthm}
\input{definitions}

\title{Knowledge Graph Enhanced Large Language Model Editing}

\author{\textbf{ Mengqi Zhang$^{1}$\footnotemark[1] , Xiaotian Ye $^{2}$\footnotemark[1] , Qiang Liu$^{3}$ , Pengjie Ren$^{1}$, Shu Wu$^{3}$ , Zhumin Chen$^{1}$} \\
$^1$School of Computer Science and Technology, Shandong University \\
$^2$School of Computer Science, Beijing University of Posts and Telecommunications \\
$^3$Center for Research on Intelligent Perception and Computing \\
  State Key Laboratory of Multimodal Artificial Intelligence Systems \\
  Institute of Automation, Chinese Academy of Sciences \\
  \texttt{\{mengqi.zhang, renpengjie, chenzhumin\}@sdu.edu.cn}\\ \texttt{yexiaotian@bupt.edu.cn} \\ \texttt{\{qiang.liu,shu.wu\}@nlpr.ia.ac.cn} \\
 } 

\begin{document}
\maketitle
\renewcommand{\thefootnote}{\fnsymbol{footnote}}
\footnotetext[1]{The first two authors contribute equally.} 
\begin{abstract}
\Acp{LLM} are pivotal in advancing \ac{NLP} tasks, yet their efficacy is hampered by inaccuracies and outdated knowledge. 
Model editing emerges as a promising solution to address these challenges. 
However, existing editing methods struggle to track and incorporate changes in knowledge associated with edits, which limits the generalization ability of post-edit \acp{LLM} in processing edited knowledge. 
To tackle these problems, we propose a novel model editing method that leverages knowledge graphs for enhancing \ac{LLM} editing, namely GLAME. 
Specifically, we first utilize a knowledge graph augmentation module to uncover associated knowledge that has changed due to editing, obtaining its internal representations within \acp{LLM}. 
This approach allows knowledge alterations within \acp{LLM} to be reflected through an external graph structure. 
Subsequently, we design a graph-based knowledge edit module to integrate structured knowledge into the model editing. 
This ensures that the updated parameters reflect not only the modifications of the edited knowledge but also the changes in other associated knowledge resulting from the editing process. 
Comprehensive experiments conducted on GPT-J and GPT-2 XL demonstrate that GLAME significantly improves the generalization capabilities of post-edit LLMs in employing edited knowledge.
\end{abstract}

\section{Introduction}
\Acfp{LLM} have achieved impressive results in various \acf{NLP} tasks due to their strong general capabilities and inherent rich world knowledge \cite{zhao2023survey}. 
However, the knowledge in \acp{LLM} may be factually incorrect or outdated \cite{cao2021knowledgeable}, thereby limiting their capabilities. 
To address these issues, model editing of \acp{LLM} has been proposed, distinguishing themselves from the traditional fine-tuning approaches. Model editing employs a more efficient and precise method to update the knowledge embedded in \acp{LLM} and has garnered widespread attention from researchers in recent years. 

Model editing primarily comprises three categories of methods: Memory-based, Meta-learning, and Locate-then-edit methods. Memory-based methods, exemplified by SERAC \cite{mitchell2022memory}, store edited knowledge in the external memory outside of LLMs, enabling the retrieval of this knowledge from memory during the inference process of \acp{LLM}. Meta-learning methods typically adopt a hyper-network to learn the weight changes for editing LLMs, such as KE \cite{de2021editing} and MEND \cite{mitchell2021fast}. To achieve more precise knowledge editing, locate-then-edit methods have been proposed. For instance, ROME \cite{meng2022locating} and MEMIT \cite{meng2022mass} directly target and update parameters corresponding to specific knowledge.

\begin{figure}
  \centering
  \includegraphics[width=\linewidth]{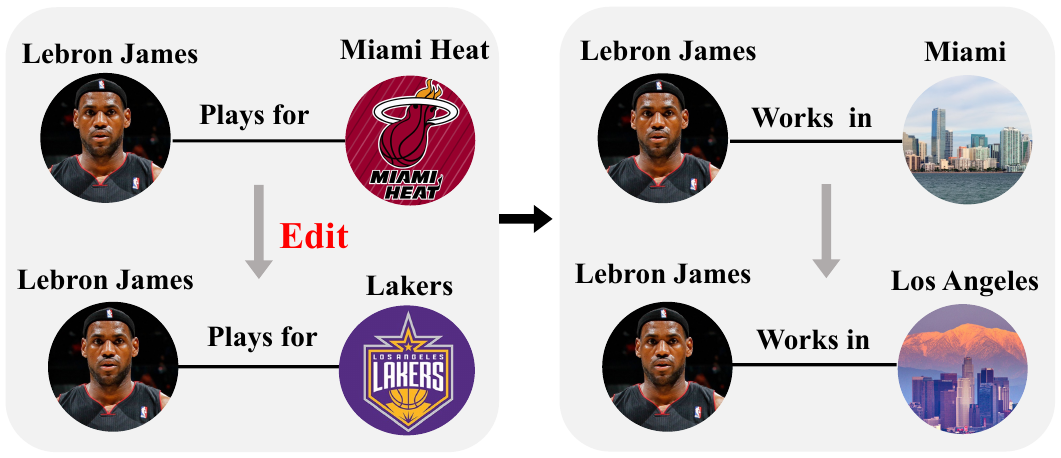}
  \caption{An example of model editing for \acp{LLM}. Editing target knowledge leads to changes in its associated knowledge.}
  \label{mov}
 \end{figure}
 
While these methods demonstrate promising results in knowledge editing of \acp{LLM}, they still face the challenge of capturing the associated knowledge changes related to edited knowledge. 
Specifically, existing work primarily focuses on the editing of target knowledge, such as modifying knowledge from $(s, r, o)$ to $(s, r, o^*)$. 
However, such single-knowledge modification often triggers a series of consequential alterations in associated knowledge. 
As shown in Figure \ref{mov}, an edit that changes the knowledge from ``\emph{LeBron James plays for the Miami Heat}'' to ``\emph{LeBron James plays for the Los Angeles Lakers}'' would necessitate a corresponding update from ``\emph{LeBron James works in Miami}'' to ``\emph{LeBron James works in Los Angeles}''. 
Existing editing methods fail to account for the impact on associated knowledge resulting from the modification of target knowledge, which limits the generalizability of post-edited \acp{LLM} in processing such edited knowledge. 
The black-box nature of \acp{LLM} makes capturing the associations between pieces of knowledge within the models exceedingly complex, further challenging the detection of such associated knowledge changes during editing.

To deal with the above challenge, we propose a novel locate-then-edit method enhanced by knowledge \underline{G}raphs for \underline{LA}rge language \underline{M}odel \underline{E}diting, namely GLAME.  
Specifically, for each target edit knowledge, we first present a knowledge graph augmentation (KGA) module (\S \ref{KGA}) to construct a subgraph that captures the new associations resulting from the edit.
Directly editing high-order relationships from the subgraph into LLMs in a simplistic way requires multiple alterations to the models and might disrupt the targeted edited knowledge, potentially exerting significant adverse effects and diminishing post-edit model performance (\S \ref{per-com}).
Therefore, we further develop a graph-based knowledge edit (GKE) module (\S \ref{GKE}) that integrates the subgraph encoding into the rank-one model editing framework. With just a single edit, it ensures that the edited parameters can recognize not only the edited knowledge but also the broader scope of knowledge impacted by such edits.

We summarize our contributions as follows:
\begin{itemize}
\item We emphasize and investigate the necessity of capturing the changes of associated knowledge induced by edited knowledge in model editing.
\item We integrate knowledge graphs into model editing and propose a novel and effective editing method to structure knowledge changes induced by editing and incorporate them into specific parameters.
\item We conduct extensive experiments on GPT-2 XL and GPT-J, which demonstrate the effectiveness of our proposed model.
\end{itemize}

\section{Related Work}
In this section, we introduce the related work on model editing, which aims to inject new knowledge into LLMs or modify their existing internal knowledge, while ensuring it does not impact other unrelated knowledge. Model editing methodologies can be broadly classified into three distinct categories \cite{yao-etal-2023-editing}: memory-based, meta-learning, and locate-then-edit approaches.

Memory-based strategies choose to augment LLMs with external memory modules, thereby offering a pathway to knowledge updates without modifying the parameters of LLMs. For example, SERAC \cite{mitchell2022memory} method introduces a gating network in conjunction with an additional model specifically designed to manage edited knowledge. However, the memory-based approaches all highlight a fundamental limitation in their scalability: the external model's management complexity escalates with each additional edit, potentially hampering its practical applicability.

Conversely, meta-learning methods eliminate the necessity for complex external memory modules by focusing on the training of a hyper-network capable of generating updated weights for the LLMs. This strategy was initially investigated by KE \cite{de2021editing}, utilizing a bi-directional LSTM to predict model weight updates. However, this approach encountered limitations when applied to larger models due to their extensive parameter spaces. To deal with this challenge, MEND \cite{mitchell2021fast} adopts a low-rank decomposition of fine-tuning gradients, showcasing an efficient mechanism for updating weights in LLMs.  Nevertheless, these approaches still require extensive computational resources for training and risk affecting unrelated knowledge.

To overcome these issues, recent works have explored knowledge location within LLMs, aiming for more interpretable and precise knowledge editing by targeting parameters directly associated with specific information. The early attempts include KN \cite{dai2022knowledge}, which proposes a knowledge attribution method to identify knowledge neurons but falls short in making precise changes to the model's weights. Subsequently, the progress in comprehending the fundamental mechanism of Transformer \cite{vaswani2017attention} models has introduced the hypothesis that the Feed Forward Network (FFN) modules might function as key-value memories \cite{geva2021transformer,geva2023dissecting}, thereby laying the groundwork for more precise editing strategies. The ROME \cite{meng2022locating} method, building on this insight, employed causal tracing to pinpoint knowledge-relevant layers and then edit its FFN module, achieving superior outcomes. Building upon this, MEMIT \cite{meng2022mass} tackles batch editing tasks, enabling large-scale knowledge integration.

Despite these advancements, all of the above models primarily concentrate on editing isolated pieces of knowledge, overlooking the potential ripple effects across the model's knowledge base \cite{cohen2023evaluating}. This omission can impair the model's generalization ability post-editing and hinder its capacity for further reasoning with newly integrated knowledge \cite{zhong2023maquake}. 

\section{Preliminaries}
In this section, we introduce the definition of model editing and knowledge graphs, and the rank-one model editing framework used in our study.

\begin{definition}[\textbf{Model Editing for LLMs}] 
Model editing \cite{yao-etal-2023-editing} aims to adjust an LLM $\mathcal{F}$'s behavior to modify the knowledge $(s, r, o)$ encoded in the model into the target knowledge $(s, r, o^*)$, where knowledge is denoted as a triple, consisting of the subject $s$, relation $r$, and object $o$. Each edit sample $e$ can be represented as $(s,r,o,o^*)$. The post-edit LLM is defined as $\mathcal{F}'$.
\end{definition}

\begin{definition}[\textbf{Knowledge Graph}]
A knowledge graph (KG) \cite{ji2021survey} stores structured knowledge as a collection of triples  $\{(s,r,o) \subseteq \mathcal{E} \times \mathcal{R} \times \mathcal{E}\}$, where $\mathcal{E}$ and $\mathcal{R}$ represent the set of entities and relations, respectively. 
\end{definition}

\subsection{Rank-one Model Editing Framework}
    Rank-one model editing (ROME) \cite{meng2022locating} is a Locate-then-edit method, this method assumes that the factual knowledge is stored in the Feedforward Neural Networks (FFNs), conceptualizing as key-value memories \cite{geva2021transformer,kobayashi2023feed}. Specifically, the output of the $l$-th layer FFN for the $i$-th token is formulated as:
    \begin{equation}
        \mathbf{m}_i^l = f(\mathbf{W}_{in}^l \cdot \mathbf{h}_i^{l-1}) \cdot \mathbf{W}^l,
    \label{fnn}
    \end{equation}
    where $f(\cdot)$ denotes the activation function, and $\mathbf{h}_i^{l-1}$ is the input of FFN. To facilitate representation, we omit the superscript $l$ in the subsequent discussion. 
    
    In this setup, the output of the first layer, $f(\mathbf{W}_{in} \cdot \mathbf{h}_i)$, serves as the keys denoted as $\mathbf{k}_i$. The outputs of the subsequent layer represent the corresponding values. Based on the hypothesis, this method utilizes casual tracing \cite{pearl2022direct,vig2020investigating} to select a specific FFN layer for editing, thereby updating the weight $\mathbf{W}$ of the second layer by solving a constrained least-squares problem:
\begin{align}
\begin{aligned}
& {\text{minimize}}
& & \|\mathbf{{W}}\mathbf{K}-\mathbf{M}\|, \\
& \text{subject to}
& & \mathbf{{W}}\mathbf{k}_* = \mathbf{m}_*.
\end{aligned} 
\end{align}
Here, the objective function aims to maintain the knowledge, irrelevant to the edited sample unchanged within the LLM, where $\mathbf{K} = [\mathbf{k}_1;\mathbf{k}_2;,\dots,;\mathbf{k}_p]$ denotes the sets of keys encoding subjects unrelated to the edited fact, and $\mathbf{M} = [\mathbf{m}_1;\mathbf{m}_2;,\dots,;\mathbf{m}_p]$ are the corresponding values. The constraint is to ensure that edited knowledge can be incorporated into the FFN layer, specifically by enabling the key $\mathbf{k}_*$ (encoding subject $s$) to retrieve the value $\mathbf{m}_*$ about the new object $o^*$.

As explicated in \cite{meng2022locating}, a closed-form solution to the above optimization problem can be derived: 
\begin{equation}
    \mathbf{\hat{W}} = \mathbf{W}+ \frac{(\mathbf{m}_*-\mathbf{W}\mathbf{k}_*)(\mathbf{C}^{-1}\mathbf{k}_*)^\mathrm{T}}{(\mathbf{C}^{-1}\mathbf{k}_*)^\mathrm{T} \mathbf{k}_*},
    \label{solution}
\end{equation} 
where $\mathbf{C}=\mathbf{K}\mathbf{K}^\mathrm{T}$ represents a constant matrix, pre-cached by estimating the uncentered covariance of $\mathbf{k}$ based on a sample of Wikipedia text (Appendix \ref{imp}). Therefore, solving the optimal parameter $\mathbf{\hat{W}}$ is transformed into calculating $\mathbf{k}_*$ and $\mathbf{m}_*$.

Extending this framework, our research delineates a method to integrate graph-structured knowledge, newly and intrinsically associated with the edited knowledge, into the editing of model parameters. We will provide a detailed description of our approach in the following sections.

\begin{figure*}
  \centering
  \includegraphics[width=1\linewidth]{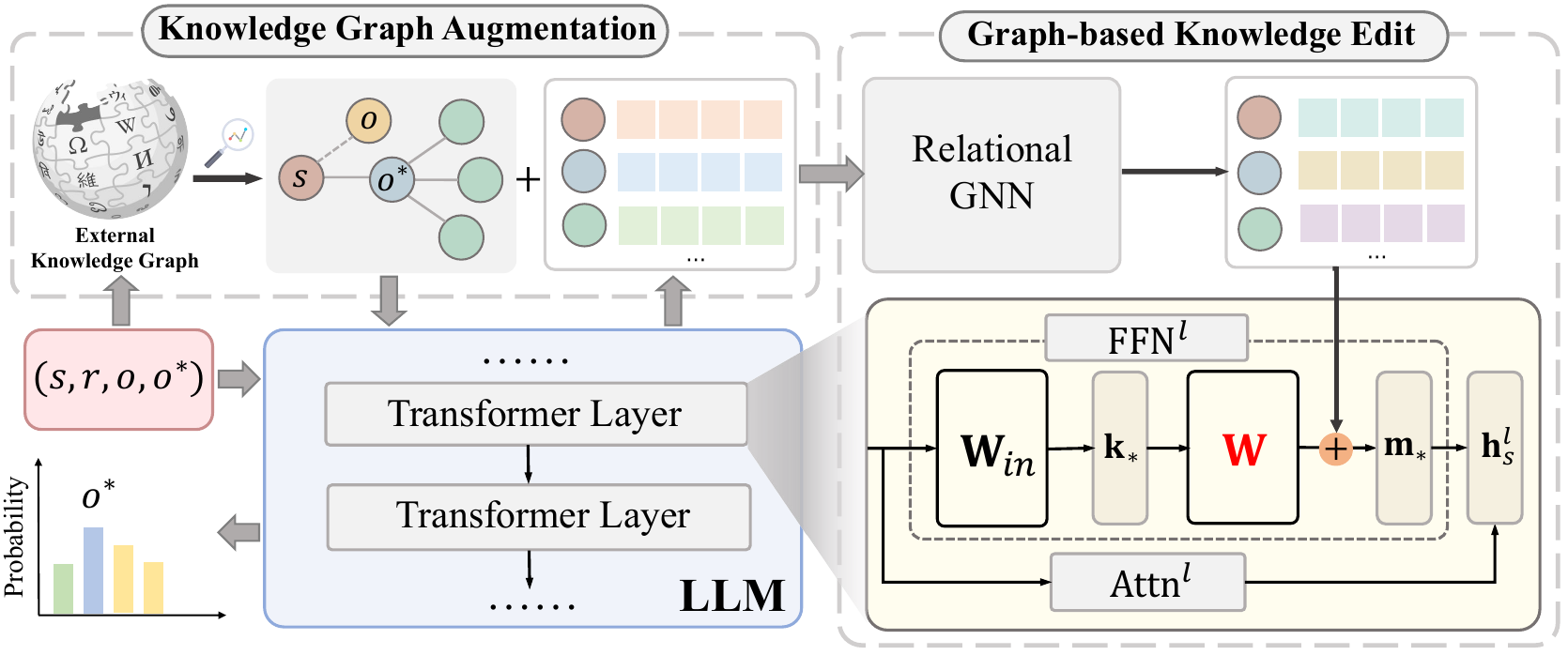}
  \caption{An illustration of GLAME architecture. We first utilize a Knowledge Graph Augmentation module to sample a high-order subgraph, recording the associated knowledge of changes caused by the edit $(s, r, o, o^*)$. Subsequently, the entities and relations within the subgraph are encoded using the LLM, from which hidden vectors are extracted from the early layers as the initial representations of the entities and relations in the subgraph. Then, the well-designed Graph-based Knowledge Edit module leverages a relational graph neural network to incorporate new knowledge associations from the subgraph into the parameter editing process. }
  \label{framework}
 \end{figure*}\textbf{}
\section{Methodology}
In this section, we introduce the proposed GLAME, the architecture of which is illustrated in Figure \ref{framework}. The \themodel framework principally comprises two key components: (1) \emph{Knowledge Graph Augmentation} (KGA), which associates the knowledge of internal changes in LLMs by utilizing external knowledge graphs, and (2) \emph{Graph-based Knowledge Edit} (GKE), which injects knowledge of edits and edit-induced changes into specific parameters of LLMs. 

\subsection{Knowledge Graph Augmentation}
\label{KGA}
To accurately capture the changes in associated knowledge induced by editing in LLMs, we propose using external knowledge graphs. This approach is divided into two operational parts: First, it leverages an external knowledge graph to construct a subgraph, capturing the altered knowledge. Then, the LLM is employed to extract the corresponding representations of entities and relations within this subgraph, serving as the initial representations. 

\subsubsection{Subgraph construction} 
\label{sub-graph}
We first introduce how to utilize an external knowledge graph to construct a subgraph that encapsulates the newly formed associations due to the edit. 

Specifically, for a given target edit sample $e = (s, r, o, o^*)$, we initially employ $o^*$ to match the most relevant entity within an external knowledge graph, such as Wikipedia\footnote{\url{https://www.wikipedia.org/}}. This step is followed by the sampling of neighboring entities and their relations centered on this entity, represented as $(o^*, r_1, o_1)$, $(o^*, r_2, o_2)$, $\cdots$, $(o^*, r_n, o_m)$. These are used to construct new two-order relationships: $(s, r, o^*, r_1, o_1)$, $(s, r, o^*, r_2, o_2)$, $\cdots$, $(s, r, o^*, r_n, o_m)$, thereby generating new associated knowledge as a consequence of editing. Here $m$ denotes the maximum number of samples for each entity.
Following this approach, we can sequentially sample the neighboring entities of $o_1$, $o_2$, $\cdots$, $o_m$, thereby constructing higher-order new knowledge associations for $s$. We define the maximum order of the newly constructed relationships as $n$. The target edit knowledge $(s, r, o^*)$, along with these new high-order relations, forms a subgraph, termed $\mathcal{G}_n^{m}(e)$, which can record changes in associated knowledge partially caused by editing knowledge. $n$ is also the maximum order of the subgraph, and together with $m$ serve as hyper-parameters to control the size of the graph. 

\subsubsection{Subgraph initialization}
\label{init-1}
To further explicitly associate the knowledge within the LLM that is affected by the edit, we extract hidden vectors of entities and relations from the early layers of LLM \cite{geva2023dissecting} as the initial representations for entities and relations in the constructed subgraph. 

In specific, we input entity and relation text into the LLM separately, and then select the hidden state vector of the last token of both the entity and the relation text in $k$-th layer as their initial representations in the subgraph:
\begin{equation}
    \label{init}
    \mathbf{z}_s, \mathbf{z}_r, \mathbf{z}_o = \mathbf{h}^k_{[s]}(s),
    \mathbf{h}^k_{[r]}(r),
    \mathbf{h}^k_{[o]}(o),
\end{equation}
where $\mathbf{h}^k_{[x]}(x)$ is the hidden state vector of the last token of text $x$ at the $k$-th layer of the LLM.

\subsection{Graph-based Knowledge Edit}
\label{GKE}
After obtaining the knowledge-enhanced subgraph, this section designs a graph-based knowledge edit module to integrate the new associated knowledge contained in the subgraph into the modified parameters of the LLM.

\subsubsection{Subgraph encoding} 
To enhance the subject $s$ with the newly constructed associated knowledge resulting from the editing of target knowledge, we perform message propagation and aggregation operations on the subgraph through a relational graph
neural network (RGNN) \cite{schlichtkrull2018modeling,10.1145/3447548.3467350}. 

Formally, we encode the subgraph as follows:
\begin{equation}
\label{sub-graph encoding}
  \mathbf{z}_{s}^{l+1} = {g}\left(\sum_{o \in \mathcal{N}_{s}} \mathbf{W}_1\left(\mathbf{z}_{o}^l+\mathbf{z}_r\right) + \mathbf{W}_2\mathbf{z}_{s}^l\right),
\end{equation}
where $\mathcal{N}_{s}$ is the set of neighbors of $s$ in $\mathcal{G}_n^{m}(e)$, $g(\cdot)$ is the ReLU function, $\mathbf{W}_1$ and $\mathbf{W}_2$ $\in \mathbb{R}^{d \times d}$ are trainable weight parameter matrices in each layer, and $\mathbf{z}_{s}^0$, $\mathbf{z}_{o}^0$, and $\mathbf{z}_r$ are the corresponding entity and relation representations obtained from \S\ref{init-1}. To capture the semantic dependencies among nodes in the subgraph comprehensively, the number of layers of RGNN is set to the subgraph's maximum order $n$, yielding the entity representation $\mathbf{z}_{s}^n$ after $n$-layer operation.

\subsubsection{Knowledge editing}
Following the ROME framework \cite{meng2022locating}, in this subsection, we target specific layer $l$ for the computation of $\mathbf{m}_*$ and $\mathbf{k}_*$. Subsequently, we employ Equation (\ref{solution}) to update the parameters of the second layer of the FNN, thereby accomplishing the editing of knowledge. 

\noindent{\bf Computing $\mathbf{m}_*$.} Given that $\mathbf{z}_s^n$ aggregates the information of neighbors under new association relationships, we utilize $\mathbf{z}_s^n$ to enhance the representation at the last token of $s$ in $l$-th FFN layer of the pre-edit LLM:
\begin{align}
\label{m}
    \mathbf{{m}}_{*}&=\mathbf{m}_s^{l} + \mathbf{z}_s^n,
\end{align}
where $\mathbf{m}_s^{l}$ denotes the output from the $l$-th FFN at the last token of $s$ in the pre-edit LLM. Further details of the FFN are delineated in Equation (\ref{fnn}). 

For each edit sample $(s,r,o,o^*)$, our objective is to refine an RGNN to produce an enhanced representation, $\mathbf{m}_*$, that enables the LLM to accurately predict the target object $o^*$. Accordingly, the primary loss function is defined as:
\begin{align*}
    \mathcal{L}_p = -&\frac{1}{N} \sum^N_{j=1}\log\mathrm{P}_{\mathcal{F}(\mathbf{m}_s^{l}:=\mathbf{m}_*)}[o^*\mid x_j \oplus p(s, r)],
\end{align*}
where $x_j$ is the random prefix generated by the LLM to foster optimization robustness. $\mathcal{F}(\mathbf{m}_s^{l}:=\mathbf{m}_*)$ indicates the LLM's inference alteration through the hidden state $\mathbf{m}_s^{l}$ modification to $\mathbf{m}_*$. 

To mitigate the impact of enhancing $s$ on its intrinsic properties within the LLM, we aim to minimize the KL divergence between $\mathcal{F}(\mathbf{m}_s^{l}:=\mathbf{m}_*)$ and the original model $\mathcal{F}$ without any interventions \cite{meng2022locating}:
\begin{align*}
    \mathcal{L}_a =D_{\text{KL}}\left(\mathrm{P}_{\mathcal{F}(\mathbf{m}_s^{l}:=\mathbf{m}_*)}[x \mid p '] \parallel \mathrm{P}_{\mathcal{F}}[x \mid p’] \right),
\end{align*}
where $p'$ denotes prompts in the form of "{subject} is a". This term serves as a regularization loss.

Ultimately, the parameters of the RGNN are optimized by minimizing the following objective function:
\begin{align}
\label{loss}
    \mathcal{L} = \mathcal{L}_p + \lambda \mathcal{L}_a,
\end{align}
where $\lambda$ adjusts the regularization strength. 
It is important to note that throughout the optimization process, the parameters of the LLM remain unchanged. The modification is instead focused on optimizing the parameters of the RGNN, which in turn influences the inference of the LLM.

\noindent{\bf Computing $\mathbf{k}_*$.} For each edit sample $(s,r,o, o^*)$, the $\mathbf{k}_*$ is calculated by 
\begin{equation}
\label{k}
    \mathbf{k}_* = \frac{1}{N}\sum_{j=1}^{N}f(\mathbf{W}_{in}^l \cdot \mathbf{h}_s^{l-1}).
\end{equation}
Here, we also utilize $N$ random prefixes generated in the same manner as for the computing $\mathbf{m}_*$  \cite{meng2022locating}.

After obtaining the optimized $ \mathbf{{m}}_*$ and $\mathbf{k}_*$, we bring them into Equation (\ref{solution}) and then get the edited parameter $\mathbf{\hat{W}}$. Algorithm \ref{alg} provides the pseudo-code of the overall framework.

\section{Experiments}
In this section, we evaluate our editing method \themodel by applying it to two datasets and assessing its performance on two auto-regressive LLMs. We aim to answer the following questions through experiments.
\begin{itemize}
\item {\bf Q1}: How does \themodel perform in editing knowledge compared with state-of-the-art model editing methods?
\item {\bf Q2}: How do different components affect the \themodel performance?
\item {\bf Q3}: How sensitive is \themodel with different hyper-parameter settings?
\end{itemize}
\subsection{Experimental Setups}
\subsubsection{Datasets and Evaluation Metrics}
We evaluate our \themodel on three representative datasets in our experiments: \textsc{CounterFact} \cite{meng2022locating}, \textsc{CounterFactPlus} \cite{yao-etal-2023-editing}, and \textsc{MQuAKE} \cite{zhong2023maquake}. 

\textbf{\textsc{CounterFact}} is a dataset that focuses on inserting counterfactual knowledge into models. 
We utilize three metrics on this dataset: \emph{Efficacy Score}, measuring the success rate of edits directly; \emph{Paraphrase Score}, indicating the model's ability to accurately recall edited knowledge in paraphrased forms, thus testing its generalization ability; and \emph{Neighborhood Score}, assessing whether irrelevant knowledge in the LLM is disturbed by testing with close, yet unrelated prompts. 

\textbf{\textsc{CounterFactPlus}}, an extension of \textsc{CounterFact}, presents more challenging test questions aimed at evaluating the post-edit models' ability to accurately respond to queries requiring reasoning with edited knowledge. Compared with \textsc{CounterFact}, this assessment has higher requirements for generalization ability. Following \cite{yao-etal-2023-editing}, we employ \emph{Portability Score} to evaluate the performance of all methods on this dataset. This metric offers a superior reflection of the models' generalization capabilities compared to other indicators.

An introduction to \textbf{\textsc{MQuAKE}}, further details on \textsc{CounterFact} and \textsc{CounterFactPlus}, as well as the evaluation metrics are shown in Appendix \ref{data} and \ref{metrics}.  We provide results on MQuAKE dataset in Appendix \ref{appendix_mquake} as an additional experiment.

\subsubsection{Baselines}
Our experiments are conducted on GPT-2 XL (1.5B) \cite{radford2019language} and GPT-J (6B) \cite{gpt-j}, and we compare \themodel with the following state-of-the-art editing methods: Constrained Fine-Tuning (FT) \cite{zhu2020modifying}, MEND \cite{mitchell2021fast}, ROME \cite{meng2022locating}, and MEMIT \cite{meng2022mass}. To further verify the superiority of our graph-based editing method, we also compare our method with two variant models ROME-KG and MEMIT-KG. These models utilize ROME and MEMIT, respectively, to directly edit the new high-order relations, $(s,r,o*,r,o_1), \cdots, (s,r,o*,r,o_n)$ constructed as described in \S\ref{sub-graph} and arising from the edited knowledge $(s,r,o,o*)$, into the LLM. We provide implementation details of baselines and \themodel in Appendix \ref{baseline}.

\begin{table*}[!t]
\centering
\begin{tabular}{cc|cccc|c}
\toprule
 &\textbf{Editor} & \textbf{Effi.Score} & \textbf{Para.Score} & \textbf{Neigh.Score} & \textbf{Port.Score} & \textbf{Edit.Score} \\
\midrule
&{GPT-2 XL (1.5B)}  & 22.20 & 24.70 & 78.10 & 10.18 & 20.35 \\
\midrule
& FT      & \textbf{100.00} & 87.90 & \textcolor{gray}{40.40} & {15.13} & 35.64  \\
& MEND    & 99.10 & \textcolor{gray}{65.40} & \textcolor{gray}{37.90} & {11.15}& 28.28 \\
& ROME    & \underline{99.95}  & \underline{96.48} & 75.44 & \underline{21.43} & \underline{49.82} \\
& ROME-KG & 73.85  & 72.41 & 74.65 & \textcolor{gray}{5.24} & 17.27 \\
& MEMIT   & 93.79  & 80.22 & \underline{77.05} & 18.71 & 44.67 \\
& MEMIT-KG & \textcolor{gray}{53.09}  & \textcolor{gray}{45.28} & \textbf{77.90} & \textcolor{gray}{9.99} & 26.00 \\
\rowcolor{gray!20}& \themodel & 99.84  & \textbf{96.62} & {76.82} & \textbf{23.95} & \textbf{53.24} \\

\midrule
&{GPT-J (6B)} & 16.30 & 18.60 & 83.00 & 11.44 & 18.64 \\
\midrule
& FT      & 100.00 & 98.80 & \textcolor{gray}{10.30} & 17.84 & 23.09 \\
& MEND    & \underline{97.40} & \textcolor{gray}{53.60} & \textcolor{gray}{53.90} & 12.99 & 32.14 \\
& ROME    & 100.00 & \underline{99.27} & 79.00 & 29.67 & 60.21 \\
& ROME-KG & 68.90  & 67.12 & 78.59 & 13.68 & 34.55 \\
& MEMIT   & 100.00 & 95.23 & 81.26 & \underline{29.77} & \underline{60.24} \\
& MEMIT-KG & \textcolor{gray}{53.75}  & \textcolor{gray}{40.22} & \textbf{82.80} & \textcolor{gray}{8.63} & 23.33 \\
\rowcolor{gray!20}& \themodel & \textbf{100.00} & \textbf{99.30} & \underline{81.39} & \textbf{33.04} & \textbf{63.87} \\
\bottomrule
\end{tabular}
\caption{Performance comparison on \textsc{Counterfact} in terms of Efficacy Score (\%), Paraphrase Score (\%), and Neighborhood Score (\%), and \textsc{CounterfactPlus} in terms of Portability Score (\%). The Editing Score (\%) is the harmonic mean of the four evaluation metrics. The best performance is highlighted in boldface, and the second-best is underlined. Gray numbers indicate a clear failure on the corresponding metric. }
\label{per}
\end{table*}
\subsection{Performance Comparison (RQ1)}
\label{per-com}
The performance of all editors on the \textsc{CounterFact} and \textsc{CounterFactPlus} is presented in Table \ref{per}. From the results, we have the following observations:

Our model \themodel secures the highest performance on the comprehensive evaluation metric, the Editing Score, surpassing other editors across most evaluation metrics. Specifically, \themodel exhibits enhancements of $11.76$ \% and $10.98$ \% in Portability Score over the best baseline models for GPT-2 XL and GPT-J, respectively. This demonstrates that our method can effectively improve the generalization ability of post-edit LLM in utilizing edited knowledge, especially in multi-hop reasoning, by effectively introducing external knowledge graphs. The editing methods based on the ROME framework, \themodel, ROME, and MEMIT, are significantly better than other methods in Paraphrase and Neighborhood Scores. The reason might be these methods impose explicit constraints on editing knowledge recall and retention of editing-irrelevant knowledge. Although MEND and FT, which directly optimize parameters, can accurately recall edited knowledge and achieve commendable results on the Efficacy Score, their lack of precision during the editing process leads to poor performance on Paraphrase, Neighborhood, and Portability Scores compared to other editors.

ROME-KG and MEMIT-KG, compared to ROME and MEMIT, demonstrate a notable degradation in performance. This indicates that simply adding extra external information for editing does not guarantee improved performance. Specifically, ROME-KG requires multiple adjustments to the model's parameters to edit high-order relationships, potentially harming the original parameters. MEMIT-KG's unconstrained incorporation of vast amounts of information into the LLM may compromise the editing of target knowledge. In contrast, \themodel, by developing an editing method tailored for graph structures, incorporates multiple pieces of associated knowledge altered due to editing into the model with just a single edit. This approach not only maintains the precision of edits but also substantially improves the efficiency of leveraging external knowledge graphs.

\begin{table*}[!t]
\centering
\begin{tabular}{cc|cccc|c}
\toprule
 &\textbf{Editor} & \textbf{Effi.Score} & \textbf{Para.Score} & \textbf{Neigh.Score} & \textbf{Port.Score} & \textbf{Edit.Score} \\
\midrule
&{GPT-2 XL (1.5B)}  & 22.20 & 24.70 & 78.10 & 10.18 & 20.35 \\
\midrule
& \themodel w/ MLP  & 99.79  & 91.79 & \textbf{77.05} & 21.73 & 50.55 \\
& \themodel w/ GNN & 99.79  & 94.95 & {77.02} & 22.59 & 51.41 \\
& \themodel w/o GKE  & \textbf{99.95}  & {96.48} & 75.44 & 21.43 & 49.82  \\
\rowcolor{gray!20}& \themodel & {99.84}  & \textbf{96.62} & {76.82} & \textbf{23.95} & \textbf{53.24} \\

\midrule
&{GPT-J (6B)} & 16.30 & 18.60 & 83.00 & 11.44 & 18.64 \\
\midrule
& \themodel w/ MLP  & 99.85  & 98.28 & 80.41 & 30.45 & 61.94 \\
& \themodel w/ GNN & 100.00  & 98.20 & 81.03 & 30.16 & 60.90 \\
& \themodel w/o GKE & 100.00 & {99.27} & 79.00 & 29.67 & 60.21 \\
\rowcolor{gray!20}& \themodel & \textbf{100.00} & \textbf{99.30} & \textbf{81.39} & \textbf{33.04} & \textbf{63.87} \\
\bottomrule
\end{tabular}
\caption{Ablation studies on \textsc{Counterfact} in terms of Efficacy Score (\%), Paraphrase Score (\%), and Neighborhood Score (\%), and \textsc{CounterfactPlus} in terms of Portability Score (\%).}
\label{var}
\end{table*}
\subsection{Ablation Studies (RQ2)}
To investigate the superiority of each component of our method, we compare \themodel with different variants: \themodel w/ GNN, which omits RGNN's relational information and employs a GNN \cite{DBLP:conf/iclr/KipfW17} for subgraph encoding in the GKE module; \themodel w/ MLP, which neglects graph structural information, relying solely on MLP for encoding entity representations within the GKE module; and \themodel w/o GKE, which removes the GKE module and degenerates into the ROME. The results are shown in Table \ref{var} and we have the following observations:

\themodel outperforms \themodel w/ MLP and \themodel w/o GKE on most evaluation metrics, especially in Portability Score and Editing Score. This confirms that integrating the structured knowledge altered due to edited samples through the GKE module can effectively enhance the generalization ability of the post-edit model. Additionally, \themodel w/ MLP and \themodel w/ GNN also achieve better performance in Editing Score than \themodel w/o GKE. The improvements verify that the effective incorporation of external information: the hidden state vector of the subject entity and its neighbors from the early layers of LLM, contributes to the performance of edits. Compared with \themodel w/ GNN, the performance of \themodel is further improved, which highlights the importance of relations in LLM's recognition of complex graph-structured knowledge associations.
 
\subsection{Sensitivity Analysis (RQ3)}
To further explore the sensitivity of \themodel to important hyper-parameters, we examine the impact of key hyperparameters, the maximum order $n$ of subgraph, and the maximum number $m$ of sampled neighbors, on the performance of \themodel. Further results are described in Appendix \ref{appendix_sens}.

\subsubsection{Effect of maximum subgraph order $n$}
\begin{figure}[t]
	\centering
	\subfloat[GPT-2 XL]{\includegraphics[scale=0.5]{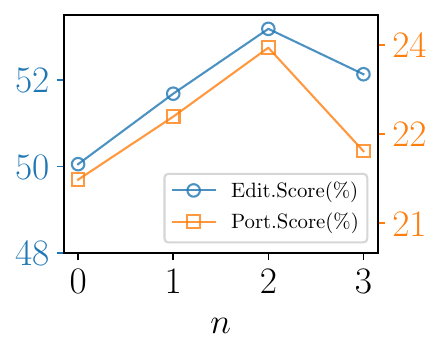}
	}
	\subfloat[GPT-J]{\includegraphics[scale=0.5]{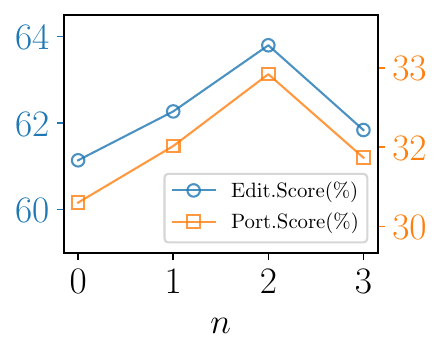}
	}
	\caption{Performance of \themodel with different subgraph order $n$ in terms of Editing and Probability Scores (the left y-axis shows Editing Score and the right y-axis shows Portability Score).
	}
	\label{order}
\end{figure}
Subgraph construction is a vital operation of the Knowledge Graph Augmentation module (\S \ref{sub-graph}). The maximum order of the subgraph decides the scope of associated knowledge affected by the edited knowledge. In this part, we conduct \themodel with different subgraph order $m$ in the GKE module on GPT-2 XL and GPT-J in terms of Editing and Portability Score. We set $m$ in the range of $\{0, 1, 2, 3\}$. The results are shown in Figure \ref{order}. The main observations are as follows: 

Increasing the maximum subgraph order $m$ significantly improves the post-edit model performance, peaking at $m=2$ for two LLMs. \themodel with $m>0$ consistently outperforms \themodel with $m=0$. We attribute the improvement to the incorporation of associated knowledge that has been altered due to editing. However, as the maximum order exceeds $2$ ($m>2$), the post-model’s performance begins to decline, which may be because the use of higher-order information makes it easy to introduce noise to the editing process.

\subsubsection{Effect of the maximum number $m$ of neighbors}
\begin{figure}[t]
	\centering
	\subfloat[GPT-2 XL]{\includegraphics[scale=0.5]{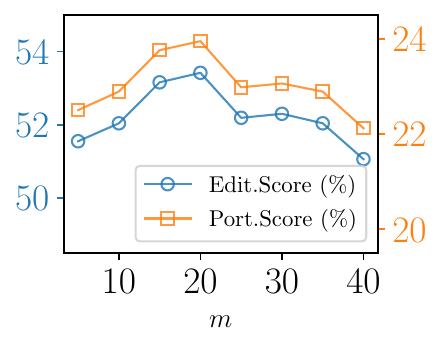}
	}
	\subfloat[GPT-J]{\includegraphics[scale=0.5]{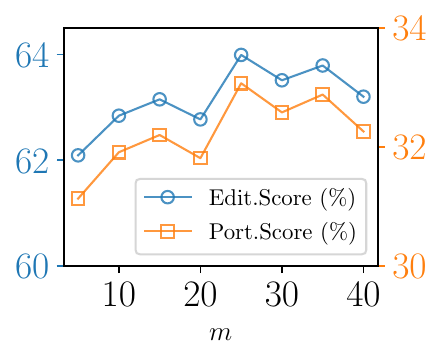}
	}
	\caption{Performance of \themodel with different maximum number $m$ of neighbors in terms of Editing and Probability Scores (the left y-axis shows Editing Score and the right y-axis shows Portability Score).
	}
	\label{size}
\end{figure}
To further investigate how the size of subgraph affects the editing performance, we conduct experiments with \themodel, varying the maximum numbers $m$ of neighbors per node within the KAG module on GPT-2 XL and GPT-J in terms of Editing and Portability Score. The results are depicted in Figure \ref{size}. Specifically, we observed a consistent improvement in editing performance as the number of neighbors increased from $5$ to $20$ for GPT-2 XL, and up to $25$ for GPT-J. This suggests that incorporating more neighbors can enhance the representation of the central entity, so that the graph structure may better reflect changes caused by edited knowledge. However, as the $n$ continued to increase, the model's performance began to decline. This decline could be attributed to the introduction of noise by an excessive number of neighboring nodes, and the increased subgraph size may escalate the optimization difficulty for the RGNN.

\section{Conclusion}
In this paper, we have proposed a novel method \themodel for large language model editing. \themodel leverages a Knowledge Graph Augmentation module to capture the changes in associated knowledge due to edit by constructing an external graph. Following this, we introduce a Graph-based Knowledge Edit module that utilizes a relational graph neural network to seamlessly integrate new knowledge associations from the constructed subgraph into the LLM's parameter editing framework. Experimental results on two LLMs and extensive analysis demonstrate the effectiveness and superiority of \themodel in model editing tasks.
\section*{Limitations}
In this section, we discuss the limitations of our \themodel. Specifically, our framework's reliance on knowledge graphs may be limited by the availability and quality of relevant knowledge. In cases where related knowledge is scarce or the knowledge graph is of low quality, the model's performance may suffer. In the future, we will develop more sophisticated subgraph sampling strategies to improve subgraph quality and more accurately capture knowledge changes resulting from editing. Additionally, these strategies aim to increase sampling speed and reduce subgraph size.

\section*{Ethical Considerations}
We realize that there are risks in developing generative \acp{LLM}, so it is necessary to pay attention to the ethical issues of \acp{LLM}.
We use publicly available pre-trained \acp{LLM}, i.e.,  GPT-2 XL (1.5B) and GPT-J (6B).
The datasets are publicly available, i.e., \textsc{CounterFact}, \textsc{CounterFactPlus}, and \textsc{MQuAKE}.
All models and datasets are carefully processed by their publishers to ensure that there are no ethical problems.

\bibliography{anthology,custom}

\appendix
\section{Pseudocode}
Algorithm \ref{alg} provides the pseudo-code of our editing method \themodel.

\SetKwComment{Comment}{/* }{ */}
\SetKwInput{KwData}{Input}
\SetKwInput{KwResult}{Output}
\SetNoFillComment
\begin{algorithm}[h!]
\caption{Editing procedure}\label{alg}
\KwData{LLM $\mathcal{F}$; Edit sample $(s,r,o,o*)$; Initial RGNN parameters}
\KwResult{The post-edit $\mathcal{F}'$}
\tcc{Subgraph Graph Construction}
Obtain subgraph $\mathcal{G}_{n}^m(e)$ from a external knowledge graph and edit sample\;
\tcc{Subgraph initialization}
$\mathbf{z}_{s},\mathbf{z}_{r},\mathbf{z}_{o}\leftarrow $ Eq (\ref{init}), $s,r,o \in \mathcal{G}_{n}^m(e)$ \;
\tcc{Optimizing $\mathbf{m}_*$}
\While{\textnormal{not converged}}{
\tcc{Subgraph encoding}
$\mathbf{z}_s^{n}\leftarrow \operatorname{RGNN}(\mathcal{G}_{n}^m(e))$ , Eq (\ref{sub-graph encoding})\;
\tcc{Computing  $\mathbf{m}_{*}$}
$\mathbf{m}_{*}\leftarrow $ Eq (\ref{m}) \;
\tcc{Learning Objective }
$\mathcal{L}\leftarrow  \mathcal{L}_p + \lambda \mathcal{L}_a$, Eq (\ref{loss})\;
Update parameters of $\operatorname{RGNN}$.
	}
\tcc{Computing $\mathbf{k}_*$}
$\mathbf{k}_*\leftarrow$ Eq (\ref{k})\;
\tcc{Updating the parameters of the FNN at the specified layer}
$\mathbf{\hat{W}}\leftarrow$ Eq (\ref{solution})\;
Return post-edit LLM $\mathcal{F}'$
\end{algorithm}

\section{Datasets Detail}
\label{data}
\subsection{Details of \textsc{CounterFact} Dataset}

\begin{table*}[t]
  \centering
  \resizebox{\linewidth}{!}
  {%
  \begin{tabularx}{\linewidth}{lX} 
    \toprule
    \textbf{Property} & \textbf{Value} \\
    \midrule
    Edit Request & The mother tongue of \{Danielle Darrieux\} is \textit{French $\to$ English}  \\
    Efficacy\_prompt & The mother tongue of Danielle Darrieux is \\
    Paraphrase\_prompt & Where Danielle Darrieux is from, people speak the language of \\
    Neighborhood\_prompt & Michel Rocard is a native speaker of \\
    \bottomrule
  \end{tabularx}
  }
  \caption{An Example of \textsc{CounterFact} Dataset}
  \label{cf_sample}
\end{table*}

Table \ref{cf_sample} shows an example from the \textsc{CounterFact} dataset. Each entry contains an edit request, several paraphrase prompts, and neighborhood prompts. In this example entry, the edit request aims to change the model's knowledge of \emph{Danielle Darrieux's mother tongue} from \emph{French} to \emph{English}. Paraphrase prompts are the semantical paraphrases of the target prompt, and neighborhood prompts are those prompts that have the same relation with the edit request but have a different subject, whose knowledge should remain unchanged by the edit. 

Our train/test dataset splits are kept the same as \cite{meng2022locating}. Similarly, we evaluate our method using the first $7500$ records on GPT-2 XL, and the first $2000$ records on GPT-J. Note that for methods not employing hypernetworks, including our \themodel, there is no requirement for training with the data from the training set.

\subsection{Details of \textsc{CounterFactPlus} Dataset}

\begin{table*}[t]
  \centering
  \resizebox{\linewidth}{!}
  {%
  \begin{tabularx}{\linewidth}{lX} 
    \toprule
    \textbf{Property} & \textbf{Value} \\
    \midrule
    Edit Request & The mother tongue of \{Spike Hughes\} is \textit{London $\to$ Philadelphia} \\
    Recalled relation & (Philadelphia, known for, cheesesteaks) \\
    New Question & What famous food is associated with the city where Spike Hughes originates from? \\
    New Answer & Cheesesteaks \\
    \bottomrule
  \end{tabularx}
  }
  \caption{An Example of the \textsc{CounterFactPlus}}
  \label{cfplus_sample}
\end{table*}

\begin{table*}[h!]
  \centering
  \resizebox{\linewidth}{!}
  {%
  \begin{tabularx}{\textwidth}{lX} 
    \toprule
    \textbf{Property} & \textbf{Value} \\
    \midrule
    Edit Request A & The type of music that \{Betty Carter\} plays is \textit{jazz $\to$ instrumental rock} \\
    Edit Request B & The name of the current head of state in \{USA\} is \textit{Donald Trump $\to$ Norodom Sihamoni} \\
    New Question & Who is the head of state of the country from which the music genre associated with Betty Carter originated? \\
    Original Answer & Donald Trump \\
    New Answer & Norodom Sihamoni \\
    \bottomrule
  \end{tabularx}
  }
  \caption{An Example of the \textsc{MQuAKE}}
  \label{mquake_sample}
\end{table*}

The \textsc{CounterFactPlus} dataset serves as a supplementary expansion of the original CounterFact dataset, selecting $1031$ entries as a subset of the original data and enriching them with new test questions based on the original content. Each entry contains the same edit request as found in \textsc{CounterFact}, with additional questions and answers that require LLM to do further reasoning based on the edited knowledge. 

An example entry from the dataset is showcased in Table \ref{cfplus_sample}. In this example entry, the edit request entails modifying the model's knowledge of \emph{Spike Hughes's mother tongue} from \emph{London} to \emph{Philadelphia}. This edit introduces new knowledge associations, such as \emph{(Spike Hughes, mother tongue, Philadelphia, known for, cheesesteaks)}, leading to a multi-hop question \emph{What famous food is associated with the city where Spike Hughes originates from?}, with the correct answer being \emph{Cheesesteaks}. The additional knowledge triple \emph{(Philadelphia, knowledge for, Cheesesteaks)} used to construct the multi-hop question is labeled as ``Recalled relation'' in the dataset.
In our work we primarily focus on the multi-hop reasoning aspect, aiming to assess \themodel's capacity to capture relevant changes in knowledge.

\subsection{Details of \textsc{MQuAKE} Dataset}
Similar to \textsc{CounterFactPlus}, \textsc{MQuAKE} is a more challenging dataset that also focuses on evaluating models' ability to perform further reasoning using newly edited knowledge. Each entry in this dataset may involve multiple edits and contain multi-hop reasoning questions that require reasoning from $2$ to $4$ hops to answer correctly, posing stricter requirements on the post-model's generalization capability. 

Table \ref{mquake_sample} illustrates an example from \textsc{MQuAKE} dataset. The example entry requires two edits to the LLM, inserting new knowledge \emph{(Betty Carter, plays, instrumental rock)} and \emph{(USA, head of state, Norodom Sihamoni)}. Accordingly, a 3-hop question ``\emph{Who is the head of state of the country from which the music genre associated with Betty Carter originated?}'' is constructed to assess the post-edit models's ability to employ edited knowledge and its associated knowledge.  Following \cite{zhong2023maquake}, our evaluation also focuses on a subset of $3000$ entries, evenly distributed across $\{2,3,4\}$-hop questions, with each category comprising $1000$ entries.

\section{Evaluation Metrics}
\label{metrics}
We adopt three widely-used metrics \cite{meng2022locating, meng2022mass}, {Efficacy Score}, {Paraphrase Score}, and {Neighborhood Score} to evaluate all editors on \textsc{CounterFact} dataset, and use {Portability Score} \cite{yao-etal-2023-editing} on \textsc{CounterFactPlus} dataset. We utilize the harmonic mean of four metrics, {Editing Score}, to evaluate each editor's overall capabilities.  Each metric is calculated as follows:

\textbf{Efficacy Score} is to test whether the post-edit LLMs can correctly recall the new target entity when given the edit prompt $p(s, r)$. It is calculated by
    \begin{equation*}
        \mathbb{E}\left[\mathbb{\mathbb { I }}\left[\mathrm{P}_{\mathcal{F}'}\left(o^{*}\mid p(s,r)\right)>\mathrm{P}_{\mathcal{F}'}\left(o\mid p(s,r)\right)\right]\right].
    \end{equation*}

\textbf{Paraphrase Score} measures the performance of the post-edit LLM on rephase prompt set ${P}^P$ of edit prompt $p(s,r)$.  The calculation is similar to the Efficacy Score:
    \begin{equation*}
    \mathbb{E}_{p \in {P}^P}\left[\mathbb{\mathbb { I }}\left[\mathrm{P}_{\mathcal{F}'}\left(o^{*}\mid p \right)>\mathrm{P}_{\mathcal{F}'}\left(o\mid p \right)\right]\right].
    \end{equation*}

\textbf{Neighborhood Score} measures whether the post-edit LLM assigns the higher probability to the correct fact on the prompt set ${P}^N$, which consists of distinct but semantically similar prompts $p(s,r)$. The calculation is defined as:
    \begin{equation*}
    \mathbb{E}_{p \in {P}^N}\left[\mathbb{\mathbb { I }}\left[\mathrm{P}_{\mathcal{F}'}\left(o^{*}\mid p \right)<\mathrm{P}_{\mathcal{F}'}\left(o\mid p \right)\right]\right].
    \end{equation*}
This metric can assess the extent of the impact that edits have on unrelated knowledge.
    
\textbf{Portability Score} measures the accuracy of the post-edit model on the multi-hop question set $P$ about the edit sample:
    \begin{equation*}
    \mathbb{E}_{p \in {P}}\left[\mathbb{I}\left[\mathcal{F}{'}(p) = o^{*}{'})\right]\right].
    \end{equation*}
Given the challenges associated with evaluating the data, the Portability Score provides a more accurate reflection of the model's generalization capabilities compared to other metrics.
\section{Baselines}
\label{baseline}
Our experiments are conducted on GPT-2 XL (1.5B) \cite{radford2019language} and GPT-J (6B) \cite{gpt-j}, and we compare \themodel with the following state-of-the-art editing methods:

{\bf Constrained Fine-Tuning (FT)} \cite{zhu2020modifying} involves fine-tuning specific layers of the LLM's parameters directly using gradient descent, while imposing a norm constraint on the weight changes to prevent catastrophic forgetting.

{\bf MEND} \cite{mitchell2021fast}  constructs a hyper-network based on the low-rank decomposition of gradients to perform editing.

{\bf ROME} \cite{meng2022locating} is based on the hypothesis that knowledge in LLMs is stored in the FFN module, and uses optimization to update a FFN layer to insert knowledge.

{\bf MEMIT}  \cite{meng2022mass} builds on the ROME method, specializing in batch-editing tasks by performing edits on a range of FFN layers.

To further verify the superiority of our graph-based editing method, we also compare our method with two variant models \textbf{ROME-KG} and \textbf{MEMIT-KG}. The two baselines aim to evaluate the performance of directly adding the same amount of external information to the LLM without using the GKE module. For each record in our test dataset, we construct edit requests that contain high-order relationships from the knowledge graph. For instance, given the original edit content \textit{"Spike Hughes originates from London $\to$ Washington"} and a related knowledge graph triple \textit{(Washington, capital of, United States of America)}, we then create a new edit request to insert this knowledge into the LLM:\textit{ "Spike Hughes originates from Washington, capital of United States of America"}, using either ROME or MEMIT.

\section{Implementation Details}
\label{imp}
We implement our \themodel method with \textbf{PyTorch}\footnote{\url{https://pytorch.org/}} \cite{Paszke:2019vf} and the \textbf{DGL}\footnote{\url{https://www.dgl.ai/}} \cite{wang2019dgl}.  Within the Knowledge Graph Augmentation (KGA) module, we set the maximum subgraph order $n$ to $2$ for both GPT-2 XL and GPT-J, with the maximum number of sampled neighbors $m$ set to $20$ for GPT-2 XL and $40$ for GPT-J. Hidden vectors for entities and relations are extracted from the $5$th layer of GPT-2 XL ($k=5$) and the $2$nd layer of GPT-J ($k=2$), respectively, to initialize the subgraph representations. For the GKE module, we perform editing operations on the $9$th layer of GPT-2 XL ($l=9$) and the $5$th layer of GPT-J ($l=5$) based on ROME's locating results. The hidden embedding sizes for the RGNN are set to $1600$ for GPT-2 XL and $4096$ for GPT-J. For RGNN optimization, the AdamW \cite{loshchilov2018decoupled} optimizer is used with a learning rate of $5 \times 10 ^{-1}$, the optimal regularization factor $\lambda$ is $6.25 \times 10^{-2}$ for \textsc{CounterFact} and $7.5 \times 10^{-2}$ for both \textsc{CounterFactPlus} and \textsc{MQuAKE}. To prevent overfitting, we perform early-stop when the loss is lower than $1 \times 10^{-2}$. Since our method does not require an additional training set for training, we select important hyperparameters on the training set.
For the covariance matrix estimation $\mathbf{C}$, which represents the pre-computed keys in a layer, we directly use the results computed by ROME \cite{meng2022locating}, which is collected using $100,000$ samples of Wikitext. The number $N$ of random prefixes generated for calculating $\mathbf{m}_*$ and $\mathbf{k_*}$ is to $50$, serving as a method of data augmentation for the original edits.
For other baselines, we conduct our experiment with the code implemented by ROME \cite{meng2022locating}, and all the settings of the baselines we compare, including the hyperparameters, are consistent with \cite{meng2022locating,meng2022mass}.

 Our experiments are conducted on NVIDIA Tesla A100 (80G) and AMD EPYC 7742 CPU. Under this configuration, given the pre-prepared subgraph, \themodel requires approximately $7$ seconds to perform an edit on the GPT-J model. For comparison, ROME takes approximately $5$ seconds for a similar task.  Given the relatively small parameter size of GNNs, \themodel does not necessitate significant additional GPU memory for optimization compared to other similar locate-then-edit models; in practice, approximately 48GB of GPU memory is sufficient for updating the GPT-J model.

\subsection{Wikidata Sampling Details}
In the Knowledge Graph Augmentation (KGA) module, we leverage Wikidata\footnote{\url{https://www.wikidata.org/}} as an external knowledge graph to construct a subgraph for each edit sample $(s,r,o, o^{*})$.
Specifically, we employ Wikidata's API\footnote{\url{https://query.wikidata.org/sparql}} to perform a SPARQL query, retrieving all outgoing edges of the entity $o*$. 
After retrieving these edges, we prioritize the triples by sorting them to foreground the most potentially valuable information. This prioritization is based on the frequency of each relation's occurrence across the dataset. Relations that appear less frequently are deemed more valuable as they may embody information of higher specificity or rarity, similar to principles of information entropy where less frequent occurrences convey more information.

As datasets \textsc{CounterFact}, \textsc{CounterFactPlus}, and \textsc{MQuAKE} are directly constructed using Wikidata, each edited entity within these datasets is linked with its corresponding Wikidata item ID, allowing for precise sampling. 
\textbf{Note that in our experiments, the constructed subgraphs are filtered to exclude the standard answers to the multi-hop questions.} This operation ensures that the improvement in model performance is attributed to an enhancement in the generalization ability, rather than simply being influenced by specific answer patterns within the subgraphs.

\subsection{Evaluation Details}
In our experiments, we assessed the Efficacy Score, Paraphrase Score, and Neighborhood Score on the \textsc{CounterFact} dataset following the method in \cite{meng2022locating}. We used specific prompts as inputs to the LLM and examined the model’s prediction probabilities for both the original entity $o$ and the edited entity $o^*$.  For the \textsc{CounterFactPlus} dataset, our assessment of the Portability Score involved prompting the LLM with multi-hop questions, and then verifying whether the output generated includes the correct answers. To accommodate variations in phrasing or synonyms between the model's output and the standard answer, fuzzy matching was employed. In practice, we utilized the partial ratio algorithm from Fuzzywuzzy\footnote{\url{https://github.com/seatgeek/fuzzywuzzy}} library, which calculates similarity based on the Levenshtein distance. Regarding the \textsc{MQuAKE} dataset, we adopt the Efficacy Score to evaluate the effectiveness of different editing methods. 

\section{Results on \textsc{MQuAKE}}
\label{appendix_mquake}
To further demonstrate the capability of \themodel in capturing the associated knowledge changes due to edits, we compare our \themodel with two competitive baseline models, ROME and MEMIT, on the more challenging \textsc{MQuAKE} \cite{zhong2023maquake} dataset. The results are shown in Table \ref{mquake_performance}. From the results, we find that our \themodel achieves significant improvements over ROME and MEMIT across questions of varying hops. With an increase in the number of hops, which necessitates a greater utilization of edited knowledge, the performance of all editing methods begins to decline. However, \themodel exhibits the highest relative improvement on $4$-hop questions than SOTA methods, which is likely attributed to our model's effective capture of associative knowledge, enabling it to construct a more solid knowledge representation. Such an advantage becomes significant in the context of $4$-hop questions, where the complexity of reasoning is markedly higher. This emphatically validates the effectiveness of our model in improving the post-edit model's generalization capacity in processing edited knowledge.

\begin{table}[t!]
\centering
\resizebox{\linewidth}{!}{%
\begin{tabular}{c|c|cccc}
\toprule
{\textbf{Editor}} &{\textbf{Average Score}} & {\textbf{2-hops}} & {\textbf{3-hops}} & {\textbf{4-hops}} \\
\midrule
GPT-2 XL (1.5B)  & 21.29  & 25.13 & 23.3 & 15.43 \\
\midrule
ROME & 29.70  & 39.80  & 31.07  & 18.23 \\
MEMIT  & 26.52  & 35.87  & 27.70 & 16.00 \\
\rowcolor{gray!20} \themodel  & \textbf{31.48}  & \textbf{41.83}   & \textbf{32.10}  & \textbf{20.50} \\
\rowcolor{gray!20} $\Delta Improve$ & {5.98\%}  & {5.10\%} & {3.32\%}  & {12.45\%} \\
\midrule
GPT-J (6B) & 16.83  & 15.80  & 23.60 & 11.10 \\
\midrule
ROME   & 33.15 &42.80 & 38.37 & 18.27 \\
MEMIT   & 27.46  & 35.77  & 33.03 & 13.57 \\
\rowcolor{gray!20} \themodel  & \textbf{35.11}  & \textbf{44.13}  & \textbf{39.87} & \textbf{21.33} \\
\rowcolor{gray!20} $\Delta Improve$ & {5.92\%}  & {3.11\%}  & {3.91\%} & {16.75\%} \\
\bottomrule
\end{tabular}%
}
\caption{Performance comparison of editors on multi-hop questions of \textsc{MQuAKE} dataset in terms of Efficacy Score (\%). }
\label{mquake_performance}
\end{table}

\section{Sensitivity Analysis}
\label{appendix_sens}
\begin{figure}[t]
	\centering
	\subfloat[GPT-2 XL]{\includegraphics[scale=0.5]{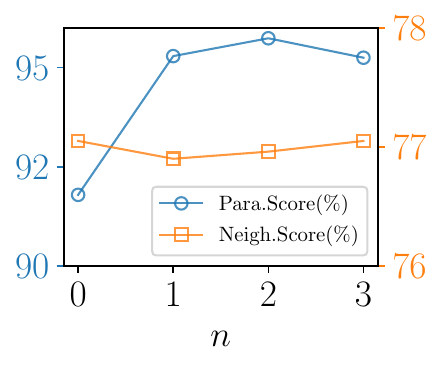}
	}
	\subfloat[GPT-J]{\includegraphics[scale=0.5]{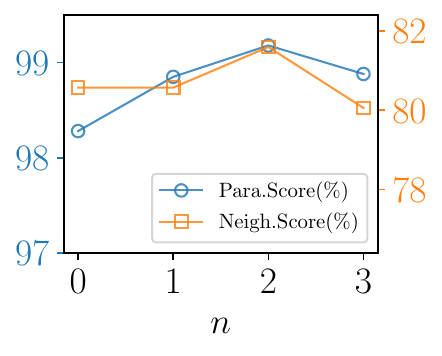}
	}
	\caption{Performance of \themodel with different subgraph order $n$ in terms of Paraphrase and Neighborhood Scores (the left y-axis shows Paraphrase Score and the right y-axis shows Neighborhood Score).
	}
	\label{order-ps}
\end{figure}

\begin{figure}[t!]
	\centering
	\subfloat[GPT-2 XL]{\includegraphics[scale=0.5]{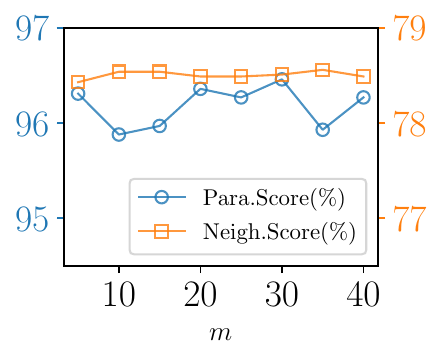}
	}
	\subfloat[GPT-J]{\includegraphics[scale=0.5]{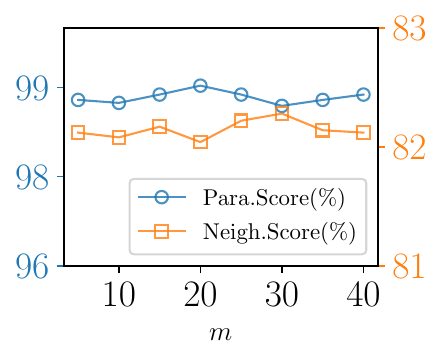}
	}
	\caption{Performance of \themodel with different maximum number $m$ of neighbors in terms of Paraphrase and Neighborhood Scores (the left y-axis shows Paraphrase Score and the right y-axis shows Neighborhood Score).
	}
	\label{size-ps}
\end{figure}

The maximum order of subgraph $n$ and the maximum number $m$ of sampled neighbors are two key hyper-parameters in \themodel. Figure \ref{order-ps} and \ref{size-ps} depict the performance of \themodel across various $n$ and $m$ values, as measured by Paraphrase and Neighborhood Score. From Figure \ref{order-ps}, we observe that increasing the order of the subgraph can enhance the post-edit model's performance in terms of the Paraphrase Score. This demonstrates that incorporating more new associated knowledge with edits can improve the generalization ability of the post-edit model in processing edited knowledge. In contrast, Neighborhood Score exhibits greater stability with respect to the value of $n$, indicating that our editing method inflicts minimal harm on the model's original capabilities. In Figure \ref{size-ps}, we can find that the Paraphrase and Neighborhood Scores are more stable than the Editing and Portability Scores in Figure \ref{size}. This stability may be attributed to the design of the loss function and those random prefixes added during optimization, which impose certain constraints on scenarios related to these two metrics, resulting in more stable behavior as the subgraph changes.

\definecolor{dgreen}{RGB}{0, 176, 80}

\begin{figure*}[t]
    \centering
    \includegraphics[width=1.0\linewidth]{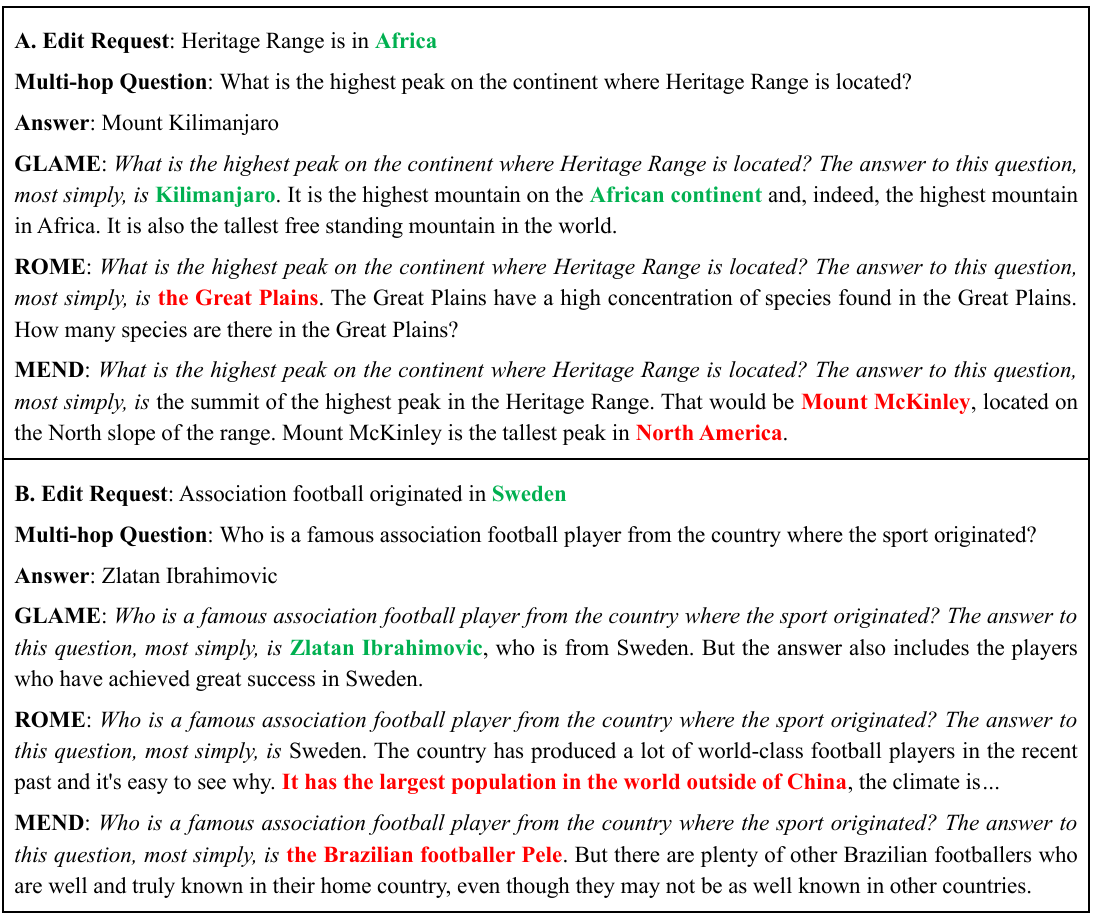}
    \caption{GPT-J generation examples of \themodel, ROME and MEND. Prompts are \textit{italic} and \textcolor{dgreen}{\textbf{green}} parts in the generation outputs are related to the multi-hop answers. \textcolor{red}{\textbf{Red}} highlights in the output indicate noticeable inconsistencies between the model-generated content and the inserted knowledge or context.}
    \label{case_study_fig}
\end{figure*}

\section{Case Study}
In this section, we present several generation examples on GPT-J utilizing three knowledge editing models: \themodel, ROME, and MEND, to demonstrate the efficacy of knowledge editing through multi-hop questions in \textsc{CounterFactPlus}. We focus on the edited models' ability to leverage newly inserted knowledge for reasoning in response to a given prompt while maintaining contextual coherence. The generation examples are shown in Figure \ref{case_study_fig}. 

\textbf{Example A [Case 1662 in \textsc{CounterFactPlus}].} In this example, counterfactual knowledge ``\emph{Heritage Range is in Africa}'' was inserted. To answer the multi-hop question correctly, the edited model must first recall the newly inserted knowledge \emph{(Heritage Range, located in, Africa)},  followed by \emph{(Africa, highest peak, Mount Kilimanjaro)}. Notably, \themodel provided the correct answer, whereas ROME and MEND seemed to fail in recalling the inserted knowledge during reasoning, offering answers such as ``\emph{the Great Plains}'' and ``\emph{Mount McKinley}'' based on Americas-related knowledge, indicating a weaker generalization.

\textbf{Example B [Case 5431 in \textsc{CounterFactPlus}].} In this example, a piece of new knowledge ``\emph{Association football originated in Sweden}'' was inserted. Answering the multi-hop question required further reasoning to identify Sweden's famous athlete, \emph{Zlatan Ibrahimovic}. \themodel maintained coherence with the context and correctly recalled the answer. Although ROME managed to recall information related to ``\emph{Sweden}'', its answer was inconsistent with the prompt, only mentioning ``\emph{Sweden}'' and mistakenly claiming ``\emph{Sweden}'' has the largest population in the world outside of China, showing signs of hallucination. MEND, again, failed to recall the newly inserted knowledge, providing an unrelated answer about the Brazilian footballer Pele.

\end{document}

%% file: definitions.tex
\theoremstyle{definition}
\newtheorem{definition}{Definition}

\acrodef{LLM}{large language model}
\acrodef{NLP}{natural language processing}
\acrodef{GLAME}{graphs for large language model editing}

\newcommand{\themodel}{\ac{GLAME}\xspace}